\documentclass[review]{elsarticle}

\usepackage{lineno,hyperref}
\modulolinenumbers[5]

\journal{Journal of \LaTeX\ Templates}

\usepackage{graphicx} 
\usepackage{amsmath}
\usepackage{algorithm}
\usepackage{algpseudocode}
\usepackage{float}
\usepackage{caption}
\usepackage{subfig}
\usepackage{color}
\usepackage{multirow}
\usepackage{makecell}
\usepackage{booktabs}

\begin{document}

\begin{frontmatter}

\title{Finite Meta-Dynamic Neurons in Spiking Neural Networks for Spatio-temporal Learning}

\author{Xiang Cheng$^{1,2}$, Tielin Zhang$^{1,2}$, Shuncheng Jia$^{1,2}$ and Bo Xu$^{1,2,3}$\corref{mycorrespondingauthor}}
\address{$^1$ Institute of Automation, Chinese Academy of Sciences (CAS), Beijing, China\\
$^2$ University of the Chinese Academy of Sciences, Beijing, China\\
$^3$ Center for Excellence in Brain Science and Intelligence Technology, CAS, China
}
\cortext[mycorrespondingauthor]{Xiang Cheng and Tielin Zhang are co-first authors of this paper. The corresponding authors are Tielin Zhang and Bo Xu. Emails: tielin.zhang@ia.ac.cn, xubo@ia.ac.cn.}

\begin{abstract}
Spiking Neural Networks (SNNs) have incorporated more biologically-plausible structures and learning principles, hence are playing critical roles in bridging the gap between artificial and natural neural networks. The spikes are the sparse signals describing the above-threshold event-based firing and under-threshold dynamic computation of membrane potentials, which give us an alternative uniformed and efficient way on both information representation and computation. Inspired from the biological network, where a finite number of meta neurons integrated together for various of cognitive functions, we proposed and constructed Meta-Dynamic Neurons (MDN) to improve SNNs for a better network generalization during spatio-temporal learning. The MDNs are designed with basic neuronal dynamics containing 1st-order and 2nd-order dynamics of membrane potentials, including the spatial and temporal meta types supported by some hyper-parameters. The MDNs generated from a spatial (MNIST) and a temporal (TIDigits) datasets first, and then extended to various other different spatio-temporal tasks (including Fashion-MNIST, NETtalk, Cifar-10, TIMIT and N-MNIST). The comparable accuracy was reached compared to other SOTA SNN algorithms, and a better generalization was also achieved by SNNs using MDNs than that without using MDNs.
\end{abstract}


\end{frontmatter}


\section{Introduction}

Many efforts have been taken towards improving artificial neural networks (ANNs) with the higher efficiency, the stronger generalization and the clearer interpretability \cite{RN699, RN668}. The biological network has played important roles in this procedure, including for example, the inspiration for the proposal of the first generation of ANNs (Perceptron), and also the following progresses after that from the perspectives of structural designments and tuning methods, towards human-competable efficient-learning such as robust computation and learing with reasoning, instead of on the contrary, such as easily fooled \cite{RN740} or attacked \cite{RN789}.

Until now, the most growing-popular type of ANNs is Deep Neural Networks (DNNs) \cite{RN660}, containing various delicately designed structures and efficient learning methods of end-to-end gradient back propagation (BP) \cite{RN726}. However, the DNN is considered as a black box, having a weak extendibility especially on tasks related to the incremental learning, and also being short on the open-ended inference and the debuggability. Some new models have been proposed to resolve these mentioned problems, such as capsule network for a better interpretability \cite{RN405}, neuronal turing machine for the biologically-plausible memorization \cite{RN378}, networks that overcomed catastrophic forgotting by reusing synapses \cite{RN788}, networks using unsupervised calsually reasoning for the robust computation \cite{RN515}.

However, these models are just at a very begining of the refinement of DNNs towards the computation with human-level intelligence. And there are still many shortcomes remained as some characteristics of DNNs that might be the main reasons for the previously described criticisms. First, the inner neuronal dynamics in DNNs is relative simple, only using activation functions describing a non-linear information conversion, e.g., Sigmoid, Tanh or Rectified Linear Unit (ReLU). Second, to the opposite, the network topologies in DNNs are surprisingly too complicated, usually with very dense connnections instead of that with sparse ones in their counterpart biological systems. Third, it is hard for the network to get a very high accuracy (that needs a little overfitting) and a strong extendability (where a little underfitting is better) at the same time. Fourth, the BP is not biologically-plausible, which makes it hard to explain the inner dynamics of the network, with the state-of-the-art biological discoveries that might be the key to open the black box of the intelligent brain. For example, the spike-timing-dependent plasticity (STDP) \cite{RN691, RN674} that described the synaptic modifications updated with spike states of pre- and postsynaptic neurons. Other local-scale plasticity rules have also been verified useful during SNN learning, including but not limited as short-term plasticity (STP) \cite{RN693}, long-term potentiation (LTP) \cite{RN684}, long-term depression (LTD) \cite{RN701} and local inhibition \cite{RN468}.

Hence, an alternative effort to break these dilemmas of DNNs is turning to the natural or spiking neural network (SNN) \cite{RN704, RN679,RN682}, which is proposed for the goal of human-like first and then human-level intelligence with both biologically-plausible structures and biologically-powerful functions. SNNs contain diverse types of biological neurons, structures, and plasticity rules that might give us more hints and inspirations for the efficient computation of biological networks from different scales.

The research on the basic biological neurons is the first step to achieve these goals. In neuroscience, different types of neurons can be classified with molecules, transcriptomes, genomes, biophysics, or morphologies \cite{RN618,RN588}. However, until now, there is still no broadly and generally agreed-upon approach for the neuron type definition and classification. Here we select the biophysics as the main neuron-type definition for its stronger relationship with neuronal dynamics. The dynamic neurons with limited hyperparameters include leaky integrated-and-fire (LIF) neuron, spike-response model (SRM) neuron, and Izhikevich neuron \cite{RN745}, describing some basic dynamics such as fast-spiking, regular-spiking, chattering-firing.

Inspired from biological neurons in the natural neural network, in this paper, we proposed a finite number of meta neurons, called Meta-Dynamic Neurons (MDN), to improve SNNs for a better network efficiency and generalization. The MDNs have non-differential membrane potentials, discrete spikes with a precise encoding of time, most importantly, the 1st-order or higher-order neuronal dynamics. Firstly, the spatial and temporal MDNs are self-learned from a spatial (i.e., MNIST) and a temporal (i.e., TIDigits) data sets, respectively. Then they are applied on other different types of tasks (e.g., Fashion-MNIST, NETtalk, Cifar-10, TIMIT and N-MNIST). The comparable accuracy was reached compared to other SOTA SNN algorithms, and a better generalization was also achieved by SNNs using MDNs than that without using MDNs.

The intrisic nature of the intelligent computation of the brain is still a mystery. However, hunderds of years' development of neuroscience might give us more hints or inspirations about it from different scales. We might start to open this black box step-by-step with the brain-inspired computation. Our research related to meta neurons is a small step forward to the better explaination of the strong generatlization of brain at micro scale, and might also contribute to the next-step research on meta curcuits at a higher mesoscale or macroscale levels.

\section{Related works}

In order to improve the generalization of DNNs, many efforts have been taken, including hyperparameter optimization, meta learning, neural architecture search, and other auto-machine-learning methods used to apply on challengable transfer-learning tasks \cite{RN426}. However, most of these efforts are still based on architectures containing only artificial neurons with a couple of simple active functions. It has been found out that the performance of DNNs could be further improved by incorporating more neuronal dynamics \cite{RN732}. Hence, in this paper, we will more focus on the biologically-plausible SNN architectures for their more complicated neuronal dynamics at the microscale.

For neuron types, the first mathematical model for biological neurons was the Hodgkin-Huxley (H-H) model \cite{RN819} described the dynamic membrane potential and a zoo of ion channels. The nonlinear LIF model is a formal threshold model of neuronal firing, which can be considered as an approximation of H-H model that contains only one attractor of membrane potential, and has been well used on many SNN algorithms \cite{RN501,RN697,RN725,RN705,RN689,RN711,RN723}. Different with H-H and LIF models that reset the membrane potential after reaching firing thresholds (also cause the non-differential problem during BP), the SRM is proposed \cite{RN820,RN821} by replacing integration of membrane potential with kernel functions. The Izhikevich neurons with 2nd-order equations of membrane potential were further proposed,  with a better presentation of complex neuronal dynamics such as chattering-firing, bursting firing, and firing with adaptation \cite{RN745,RN237}.

For plasticity rules, unsupervised local-plasticity STDP was used for the network tuning of multi-layer SNNs containing LIF neurons \cite{RN689,RN696,RN641,RN691}. The non-local reward signal was also given for the training of some hidden layers in SNNs \cite{RN694}. The balanced tuning methods were also proposed for the efficient learning of SNNs \cite{RN697,RN673}. Other biologically-plausible plasticity rules including LTP, LTD, short-term plasticity, local inhibition were also proposed \cite{RN705,RN691, RN674,RN693}. Some BP-related methods were proposed, such as BP-like STDP \cite{RN641,RN711}, spatial-temporal BP \cite{RN643}. Usually, the non-differential characteristic of SNNs would make it harder to be directly tuned with the standard BP. However, some researchers got around of this problem by tuning ANNs first and then converting them into SNNs \cite{RN686,diehl2015fast}. The main reason for this successful conversion is that the neurons in SNNs might be designed as simple LIF neurons that usually show a linear relationship between the output fire rate and the input strength of stimulus, so as to approximately convert from firerate to spikes. Other similar efforts replaced the non-differential parts of BP as a constant differential variable \cite{RN680} for an efficient learning \cite{RN732,RN680}, hence they were also called pseudo BP.

For structures, some standard modules in DNNs were also introduced into SNNs, such as convolutional kernels\cite{RN696,RN694}, feed-forward propagation \cite{RN673} and recurrent loops \cite{zhang2015a,RN822}. The performances of SNNs on benchmark datasets are increasing dramatically, including spatial datasets (e.g., MNIST, Fashion-MNIST, Cifar-10), temporal datasets (e.g., TIDdigits and TIMIT), and hybrid datasets (e.g., N-MNIST). In these benchmark datasets, a balance-tuning method was proposed for the SNN learning, which made an additional constrain of each neuron for the input-output signal balancing and got 98.6\% accuracy on the MNIST dataset \cite{RN723, RN673}. A SOM-SNN is proposed with the integration of firerate and spikes, and got 97.40\% accuracy on the TIDigits \cite{wu2018a} dataset. A three-layer SNN is proposed, tuned with curiosity mechanism and STDP learnign rule, and got 52.85\% accuracy on the Cifar-10 dataset \cite{RN762}.  Inspired by the curiosity-based learning of the human brain, in which the fancy visual sensations were more easily to be strengthened as the future memory compared to the usual stimulus, the curiosity-based SNN was proposed for the efficient learning of previous benchmark datasets with relatively fewer training samples \cite{RN725}, including MNIST and Cifar-10 datasets.

Until now, the meta learning on SNNs is only at the begining of research. In this paper, we think some meta neuronal dynamics in biological networks will improve SNNs on the efficient learning, especially on some related and new tasks.

\section{Method}

\subsection{The architecture of SNN}

In this paper, we focus more on basic neuronal dynamics instead of network topologies. Hence, a basic three-layer SNN as simple as possible is constructed for the next-step analyses. This designment minimalize the influence of complex structures to the representation of inner dynamics of neurons.

\begin{figure}[htpb]
\centering
\includegraphics[width=8.8cm]{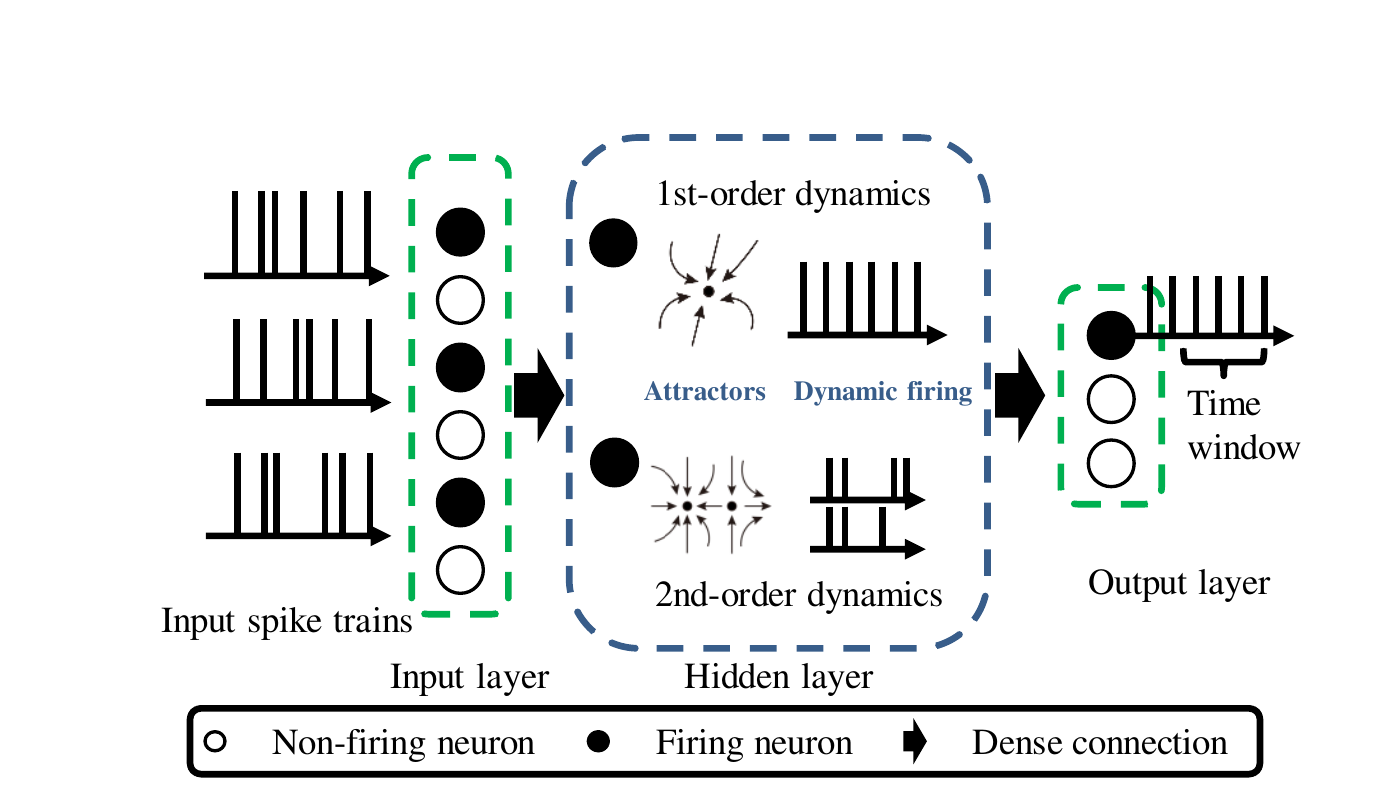}
\caption{The architecture of SNN with meta dynamic neurons.}
\label{fig_networkstructure}
\end{figure}

The architecture of SNN with MDNs is shown in Fig. \ref{fig_networkstructure}. The inputs of the network are characterized with temporal encoding of signals. The signals in hidden and output layers of the network are all spike trains with the same time span. Average firing rate is only used for the calculation of loss function and accuracy after output during learning. The MDNs contain 2nd-order dynamic membrane potentials with up to two attractors or 1st-order dynamic membrane potentials with up to one attractor.

\subsection{The 2nd-order dynamic neurons in SNN}

Izhikevich neurons \cite{RN745} are neurons with standard 2nd-order dynamics, as shown in Equation (\ref{equa_Izhikevich}), where the $V_j(t)$ is membrane potential, and the dimension of it is $mV$. The proportion of $\left(V_j(t)\right)^2$ is carefully designed as $0.04$, and also for that of $\left(V_j(t)\right)^1$ (set as $5$) and $\left(V_j(t)\right)^0$ (set as $140$). $U_j(t)$ is a membrane recovery variable charging for the activation of potassum ionic current and inactivation of sodium ionic current, and with the bigger $U_j(t)$, the $V_j(t)$ is more about hyperpolarization. After firing, the dynamic $V_j(t)$ and $U_j(t)$ will be reset to $c$ and $U_j(t)+d$, respectively. The $a$, $b$, $c$ and $d$ in Equation (\ref{equa_Izhikevich}) are the predefined hyper parameters for Izhikevich neurons.

\begin{equation}
\left\{\begin{array}{l}
\frac{dV_j(t)}{dt}=0.04V_j(t)^2 + 5V_j(t) + 140 -U_j(t)+ I\\
\frac{dU_j(t)}{dt}=a (b V_j(t)-U_j(t)) \\
\begin{matrix}
V_j(t)=c, & U_j(t)=U_j(t)+d & if(V_j(t)=30 mV)
\end{matrix}
\end{array}\right.
\label{equa_Izhikevich}
\end{equation}

Similar but different with Izhikevich neurons, we design the 2nd-order differential equations as the basic neuron model without any predefiend parameters for $\left(V_j(t)\right)^2$, $\left(V_j(t)\right)^1$, and $\left(V_j(t)\right)^0$, as shown in Equation (\ref{equa_2order}), where $U_j(t)$ is a resistance item simulating hyperpolarization, $\theta_a$, $\theta_b$, $\theta_c$, and $\theta_d$ are dynamic parameters that discriminate different 2nd-order dynamics of membrane potential $V_j(t)$ (we choose not to use $a$, $b$, $c$ and $d$ to avoid the possible occurance of confusion). For simplicity, we use 1 instead of the basic dimensions, e.g.\ mV, ms.

\begin{equation}
\left\{\begin{array}{l}
\frac{dV_j(t)}{dt}=V_j(t)^2 - V_j(t) - U_j(t)+ \sigma(\sum_{i=1}^NW_{i,j}I_i(t))\\
\frac{dU_j(t)}{dt}=\theta_a (\theta_b V_j(t)-U_j(t)) \\
\begin{matrix}
V_j(t)=\theta_c, & U_j(t)=U_j(t)+\theta_d & if(V_j(t)>V_{th}) \\
\end{matrix} \\
S_j(t) = V_j(t)>V_{th}
\end{array}\right.
\label{equa_2order}
\end{equation}

The attractor of $V_j(t)$ is decided by both $U_j(t)$ and input currents $I(t)=\sigma(\sum_{i=1}^NW_{i,j}I_i(t))$, where $I_i(t)$ is the input from upstream neuron $i$, $W_{i,j}$ is the synaptic weight between presynaptic neuron $i$ and postsynaptic neuron $j$, and $\sigma()$ is a sigmoid function used to limit the range of input currents. When $\epsilon>0$ ($\epsilon=U_j(t)-\sum_{i=1}^NW_{i,j}I_i(t)+\frac{1}{4}$, the same hereinafter), there are two extreme points of $V_j(t)$, i.e.\ $V_j^*(t)=\frac{1}{2}\pm\sqrt{\epsilon}$ which contains a fixed point attractor at $V_j^*(t)=\frac{1}{2}-\sqrt{\epsilon}$ and an unstable point at $V_j^*(t)=\frac{1}{2}+\sqrt{\epsilon}$. When $\epsilon=0$, the only extreme point of $V_j(t)$ at $V^*_j(t)=0.5$ acts as the attractor. When $\epsilon<0$, $V_j(t)$ possesses neither extreme point nor real-valued attractor, but a virtual attractor leading $V_j(t)$ to $+\infty$. Meanwhile, the existence of the attractor of $U_j(t)$ only depends on dynamic parameter $\theta_a$. When $\theta_a<0$, $U_j(t)$ has an attractor at $U_j^*(t)=bV_j(t)$ and when $\theta_a>0$, $U_j(t)$ has a virtual attractor at $+\infty$. Specially, when $\theta_a=0$, $U_j(t)$ is a constant wihout dynamic characteristic. When $V_j(t)>V_{th}$, a spike is generated as $S_j(t)=1$, or else $S_j(t)=0$.


\subsection{The procedure of generation and selection of the 2nd-order dynamic neurons}

\begin{figure}[htpb]
\centering
\includegraphics[width=7cm]{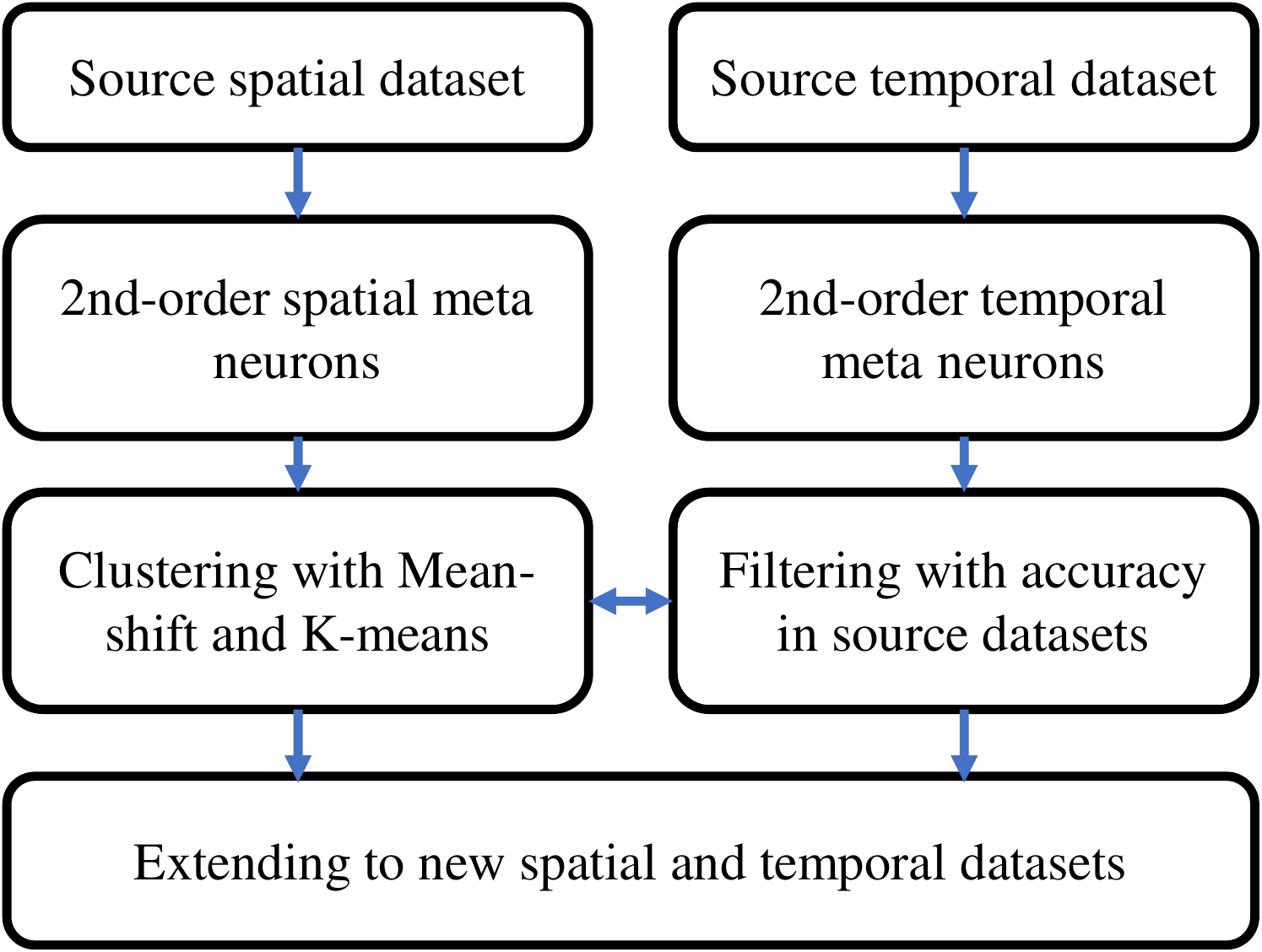}
\caption{The procedure of generating MDNs.}
\label{fig_procedure}
\end{figure}

In order to construct different kinds of 2nd-order meta neurons, we adopt a training-and-sorting approach, as shown in Fig. \ref{fig_procedure}. Firstly, we use approximate BP to train dynamic parameters of 2nd-order dynamic neurons and gain a series of candidate hyperparameters, i.e.\ $\theta_{a,k}$, $\theta_{b,k}$, $\theta_{c,k}$, and $\theta_{d,k}$, where $k\in N$ ($N$ is the number of neurons in hidden and output layers).

Secondly, we used a clustering method to process these learned hyperparameters. Since dynamic parameters $\theta_{a,k}$ and $\theta_{b,k}$ are more related to the resistance item $U$, they are combined together into a two-dimension vector while clustering (e.g.\ Fig. \ref{fig_neuron}(c,d)). Similar to that, dynamic parameters $\theta_{c,k}$ and $\theta_{d,k}$ are combined together for their naturally similar connection with the reset process (e.g.\ Fig. \ref{fig_neuron}(e,f)).

Thirdly, the number of cluster centers (labeled as $M$) is calculated by Mean-shift algorithm, and then K-means algorithm is used to accomplish the cluster process with $K=M$. The clustering centers are selected as candidate dynamic parameters to combine into complete sets of dynamic parameters.

Finally, we preliminarily sort the 2nd-order dynamic neurons according to their different dynamic characteristics of membrane potential comparing with each other. The dynamic parameters gained by this process will not be modified during the training process in next-step tasks aiming to test their convergence and generalization.

\subsection{The traditional 1st-order dynamic neurons in SNN}

In order to reveal the distinction between the 2nd-order and the 1st-order dynamic neurons, we also bring LIF to following learning tasks. LIF is a type of commonly used standard 1st-order dynamic neurons which contains only up to one attractor. The dynamic firing process at neuron $j$ can be described by Equation (\ref{equa_LIF}), where $V_j(t)$ is the membrane potential, $g_j$ is the conductance, $V_{th}$ is the predefined firing threshold, $I_i(t)$ is the input from upstream neuron $i$, $W_{i,j}$ is the synaptic weight between presynaptic neuron $i$ and postsynaptic neuron $j$, and $\sigma()$ is a sigmoid function used to limit the range of input currents.

\begin{equation}
\left\{\begin{array}{l}
\frac{dV_j(t)}{dt}=g_jV_j(t) + \sigma(\sum_{i=1}^NW_{i,j}I_i(t))\\
\begin{matrix}
V_j(t)=V_{reset} & if(V_j(t)>V_{th})
\end{matrix}\\
S_j(t) = V_j(t)>V_{th}
\end{array}\right.
\label{equa_LIF}
\end{equation}

The dynamics of membrane potential in LIF contain three processes. Firstly, before current inputs are given, $V_j(t)$ dynamically decays to the rest potential with the attractor at $V_j^*(t)=0$. Secondly, after receiving current inputs $I(t)=\sigma(\sum_{i=1}^NW_{i,j}I_i(t))$, the single attractor of membrane potential $V_j(t)$ becomes $V_j^*(t) =\frac{1}{g_j}\sum_{i=1}^NW_{i,j}I_i(t)$, which is usually bigger than $V_j(t)$ itself, thus elevating $V_j(t)$ to a higher value. Third, if $V_j(t)$ surpasses the firing threshold, that neuron fires a spike to its downstream neurons with $S_j(t)=1$, and the $V_j(t)$ will be reset as $V_{reset}$. 

For various types of LIF neurons with different hyperparameters, e.g.\ $g_j$, $V_{th}$, and $V_{reset}$, they possess different rules to update membrane potential, i.e.\ different neuronal dynamics. However, this issue of 1st-order dynamic neuron has been extensively studied and is not the concern of this paper, so we only predefine parameter $g_j=0.8$, $V_{th}=0.5$, and $V_{reset}=0$ directly (without dimensions). Additionally, we also use 1 instead of the basic dimensions for simplicity.

\subsection{Training SNN with approximate BP}

The conventional BP is based on differential chain rule, where $L$ is the standard loss function described the summation of the square error of mean output firerates and labels $O_k$, where $k$ is the id of neurons. 

\begin{equation}
L =  \sum_k^K \left(\frac{1}{T}\sum_{t}^T S_k(t)-O_k \right)^2 
\end{equation}

As shown in Equation (\ref{equa_fb_bp}), for standard BP, the $\frac{\partial L}{\partial W_{i,j}}$ contains two parts: One is the global $\frac{\partial L}{\partial V_j(t)}$ and another is local $\frac{\partial V_j(t)}{\partial W_{i,j}}$. However, $\frac{\partial L}{\partial V_j(t)}$ contains an special part $\frac{\partial S_j(t)}{\partial V_j(t)}$, which is an infinite differential gradient, where the differential of spikes is $+\infty$ or $-\infty$, hence will truncate chain rule and make SNNs untrainable by standard BP.

\begin{equation}
\left\{\begin{array}{l}
\frac{\partial L}{\partial V_j(t)}=\frac{\partial L}{\partial S_j(t)}\frac{\partial S_j(t)}{\partial V_j(t)} + \frac{\partial L}{\partial V_j(t+1)}\frac{\partial V_j(t+1)}{\partial V_j(t)}\\
\frac{\partial L}{\partial W_{i,j}}=\frac{\partial L}{\partial V_j(t)}\frac{\partial V_j(t)}{\partial W_{i,j}}
\end{array}\right.
\label{equa_fb_bp}
\end{equation}

An approximate BP trick \cite{RN732} may well resolve this problem by defining a pseudo differential gradient with lower and upper bounds for the infinite differential gradient at the spiking time within a time window. 

\begin{equation}
Grad=
\left\{\begin{array}{l}
\begin{matrix}
1 	& if(|V_j(t)- V_{th}|<V_{window}) \\
0 	& else
\end{matrix}
\end{array}\right.
\label{equa_bp}
\end{equation}

As shown in Equation (\ref{equa_bp}), $V_j(t)$ is the membrane potential saved from the phase of feed-forward propagation, $V_{th}$ is the firing threshold, $V_{window}$ is a range of membrane potential between $\delta t$, $Grad$ is the gradient calculated for the update of synaptic modifications in SNNs. This $Grad$
will be not $+\infty$ or $-\infty$, hence might propagate gradient accordingly without non-differential conflicts. 
\section{Experiments}

\subsection{The spatial and temporal datasets}

The MNIST \cite{RN657} contains 70,000 28$\times$28 1-channel gray hand-written digit-number images from 0-9, in which 60,000 samples are selected as training samples, and remained 10,000 samples are used for testing.

The Fashion-MNIST has the same number of samples and classes with MNIST dataset, only replacing the hand-written digit numbers with image classes with objects containing more complex features, such as the T-shirt and shoes.

The NETtalk \cite{RN212} is a phonetic transcription dataset, each sample in it is an English word and has multiple target phonemes. The dataset contains 5,033 training and 500 test samples with aligned English letters and phonetic representations with stresses. The input is a string of English letters with a dimension of 189 sizes (contains seven words and each word is encoded with a 27-length one-hot vector), and output is the proper phoneme to each word (a 26-length vector for 72 phonetic representations except punctuation).

The Cifar-10 \cite{RN800} contains 60,000 32$\times$32 3-channel color images covering 10 classes, in which 50,000 for training and 10,000 for testing. The samples in both of them are static 2D images.

The Neuromorphic-MNIST (N-MNIST) dataset \cite{orchard2015converting} has the same classes and samples with MNIST dataset, but after a spiking conversion by mounting the ATIS sensor on a motorized pan-tilt unit and having the sensor move while it views MNIST examples on an LCD monitor.

The TIDigits \cite{RN798} contains 4,144 (20K Hz and around 1 second for each sample) spoken digits from 0-9. It is challenging on the temporal information processing, containing sequential pronounce of spoken numbers. Signals in TIDigits are continuous; hence a further procedure of spike generation is needed. Here we set 30 frames and 30 bands with 50\% overlap for each frames, and the spikes are generated with random sampling.

The TIMIT \cite{RN823} is designed for automatic speech recognition. However, in this paper, we only use it as the dataset for gender separation. The dataset contains voices from 326 man and 136 woman for training, and lefted voices from 112 men and 56 women for testing. Each sample is a wav file with 6-bit and 16kHz, processed after MFCC with $800$ frames and $13$ bands, and then sequantially added into the SNN with timestep of $20$ (instead of direct classification with spectrogram).

Additionally, for datasets that consist of static 2D-images, e.g., MNIST, Fashion-MNIST, Cifar-10, a further procedure of spike train generation with random sampling within a time window is needed.

\subsection{The configurations and hyperparameter settings for following experiments.}

In this paper, the experiments are implemented on NVIDIA TITAN Xp. The code is written under the PyTorch framework, thus weights are randomly initialized by the default method of PyTorch. We use Adam as an optimizer and decay learning rate by epochs. Every accuracy that we present below is the average of five repeatability tests with different random seeds.

As for hyperparameters, we focus on 2nd-order neuronal dynamics in SNNs; hence we do not chase for the best dynamic-irrelevant hyperparameters but treat them as extraneous variables in control experiments, as shown in Table \ref{tab_hyperparameter}. Additionally, the initial value of $\theta_a$, $\theta_b$, $\theta_c$, $\theta_d$, $V_j$, and $U_j$ are 0.02, 0.2, 0, 0.08, 0, and 0.08 for all following tasks, respectively.

\begin{table}[htbp]
    \centering
    \footnotesize
    \caption{Hyperparameters used in following experiments, where F-MNIST is Fashion-MNIST, $\eta_W$ is the learning rate of synaptic weights between neurons.}\label{tab_hyperparameter}
    \setlength{\tabcolsep}{0.5mm}{
    \begin{tabular}{lrrrrrrr}
    \toprule
    Parameters  & MNIST & F-MNIST & NETtalk &Cifar-10 & TIDigits & TIMIT & N-MNIST \\
    \midrule
    Input size  & 784  & 784  & 189  & 3072 & 30   & 520  & 2592 \\
    Hidden size & 500  & 500  & 500  & 1500 & 500  & 500  & 500  \\
    Output size & 10   & 10   & 26   & 10   & 10   & 2    & 10   \\
    Batch size  & 100  & 100  & 5    & 100  & 10   & 32   & 100  \\
    Epochs      & 20   & 20   & 20   & 20   & 30   & 20   & 20   \\
    $\eta_W$    & 1e-3 & 1e-3 & 1e-3 & 1e-4 & 1e-2 & 1e-3 & 1e-3  \\
    \bottomrule
    \end{tabular}}
\end{table}

\subsection{The clustering of the 2nd-order MDNs}\label{dynamic}

\begin{figure}[htpb]
\centering
\includegraphics[width=10cm]{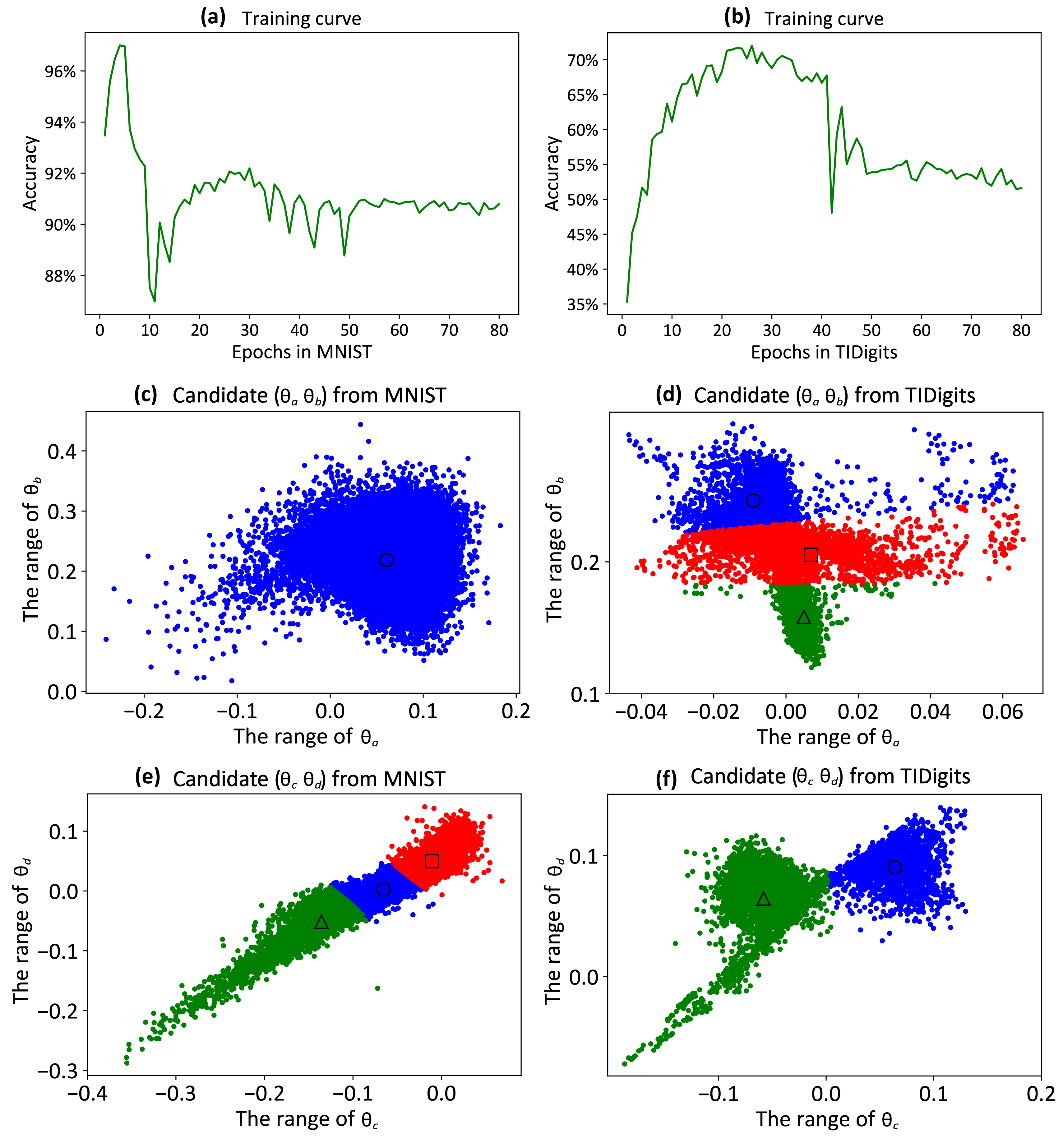}
\caption{The constructing process of dynamic parameters. Subfigures (a) and (b) are the testing accuracy curves of MNIST and TIDigits, respectively. For spatial MNIST task, the clustering centers of $\theta_a$ and $\theta_b$ are shown in subfigure (c) while that of $\theta_c$ and $\theta_d$ are shown in subfigure (e). Meanwhile, the clustering centers of $\theta_a$-$\theta_d$ for temporal TIDigits task are shown in subfigures (d) and (f).}
\label{fig_neuron}
\end{figure}

We selected spatial MNIST and temporal TIDigits as two basic source datasets for the learning of 2nd-order meta neurons and applied training of not only connection weights but also dynamic parameters of neurons with approximate BP (shown in Fig \ref{fig_neuron}(a-b)), during which the $\eta_D$ is used as the learning rates of dynamic parameters in the learning of 2nd-order meta neurons, and $\eta_D$= 1e-3 for MNIST and  $\eta_D$=1e-4 for TIDigits, respectively.

For spatial MNIST, we got only one clustering center for parameters $\theta_a$ and $\theta_b$ in Fig \ref{fig_neuron}(c), and three clustered centers for parameters $\theta_c$ and $\theta_d$ in Fig \ref{fig_neuron}(e). There were 3 types of combinations of dynamic parameters (1 center of ($\theta_a$,$\theta_b$) and 3 centers of ($\theta_c$,$\theta_d$)). These 3 2nd-order meta neurons were learned from spatial MNIST dataset hence we named them as ``spatial meta neurons''.

Similar to that, we also got another 6 ``temporal meta neurons'' (3 centers of ($\theta_a$,$\theta_b$) and 2 centers of ($\theta_c$,$\theta_d$)) from temporal TIDigits dataset, as shown in Fig. \ref{fig_neuron}(d) and Fig. \ref{fig_neuron}(f). These generated meta neurons with similar neuronal dynamics on membrane potential would further be filted out so as to get the meta neurons with both high accuracy and also larger inner differences (only one meta neuron would be left from the ones with similar dynamics of membrane potential).

\begin{figure*}[htbp]
\centering
\includegraphics[width=12cm]{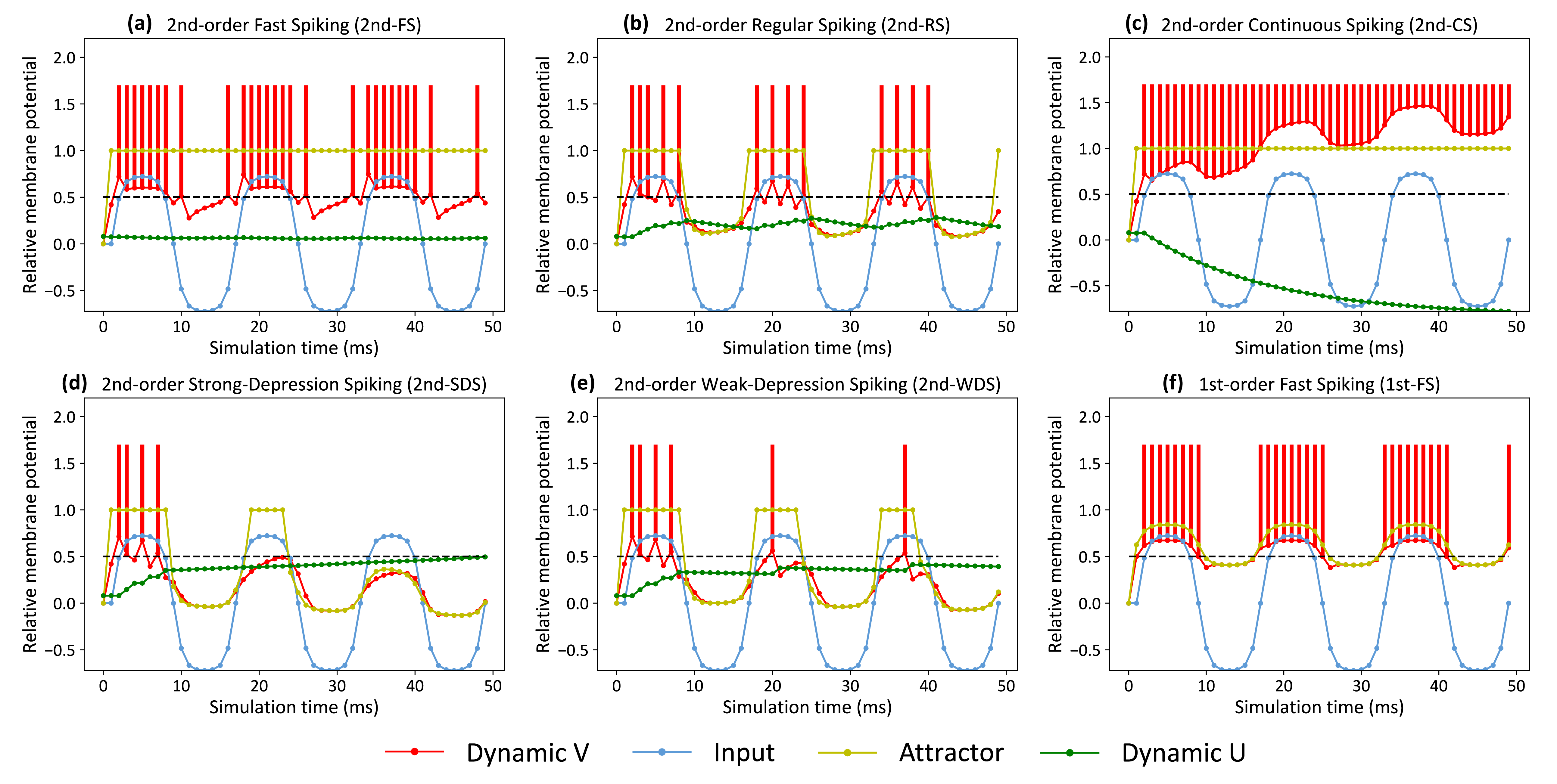}
\caption{The neuronal dynamics on membrane potential of five 2nd-order and one 1st-order meta neurons. (a-c), The spatial 2nd-order meta neurons, showing dynamic membrane potential $V$ (red lines), dynamic resistance item $U$ (green lines), simulated input (blue lines), and the attractors of membrane potential (yellow lines). (d-e), The temporal 2nd-order meta neurons, given the same input with (a-c). (f), The 1st-order meta neuron constructed from the standard LIF neuron model, given the same input with (a-c).}
\label{fig_dynamics}
\end{figure*}


\subsection{The filtering of 2nd-order meta neurons with memrbane-potential dynamics}

Firstly, dynamic analyses on membrane potential were applied on these meta neurons. As shown in Fig. \ref{fig_dynamics}(a-e), 3 spatial and 2 temporal 2nd-order meta neurons were selected from total 9 neurons (containing 3 spatial and 6 temporal neurons, as described previously) and stimulated by the same periodically fluctuating input (sine waves with the mean of 0 and the standard deviation of 0.723, i.e.\ blue lines in the subfigures of Fig. \ref{fig_dynamics}). The membrane potential $V_j(t)$ (red lines) dynamically changed according with the input stimulus. The inner variable $U_j(t)$ (green lines) also dynamically changed. The attractor of membrane potential (yellow lines) was calculated based on Equation (\ref{equa_2order}) and Equation (\ref{equa_LIF}). When real-valued attractor was not existed in 2nd-order meta neuron, which meant the membrane potential tended to elevate bondlessly, with a voltage much higher than the threshold was used to represent it.

\begin{table}[htbp]
    \centering
    \caption{Dynamic parameters used in experiments.}\label{tab_parameter}
    \begin{tabular}{lrrrr}
    \toprule
    Parameters       & 2-FS   & 2-RS   & 2-SDS  & 2-WDS  \\
    \midrule
    Param $\theta_a$ & 0.060  & 0.060  & -0.009 & 0.005  \\
    Param $\theta_b$ & 0.219  & 0.219  & 0.246  & 0.158  \\
    Param $\theta_c$ & -0.065 & -0.010 & -0.058 & -0.058 \\
    Param $\theta_d$ & 0.003  & 0.050  & 0.065  & 0.065  \\
    \bottomrule
    \end{tabular}
\end{table}

It was obviously to find that the 2nd-order Fast Spiking (2nd-FS, Fig \ref{fig_dynamics}(a)) neuron and Regular Spiking (2nd-RS, Fig \ref{fig_dynamics}(b)) neuron showed relatively strong and weak linear relationships between output firerates and input strength, respectively, which, to some extent, were similar to that of 1st-order meta neuron under the same input stimulus (Fig \ref{fig_dynamics}(f)).

\begin{figure}[htpb]
\centering
\includegraphics[width=10cm]{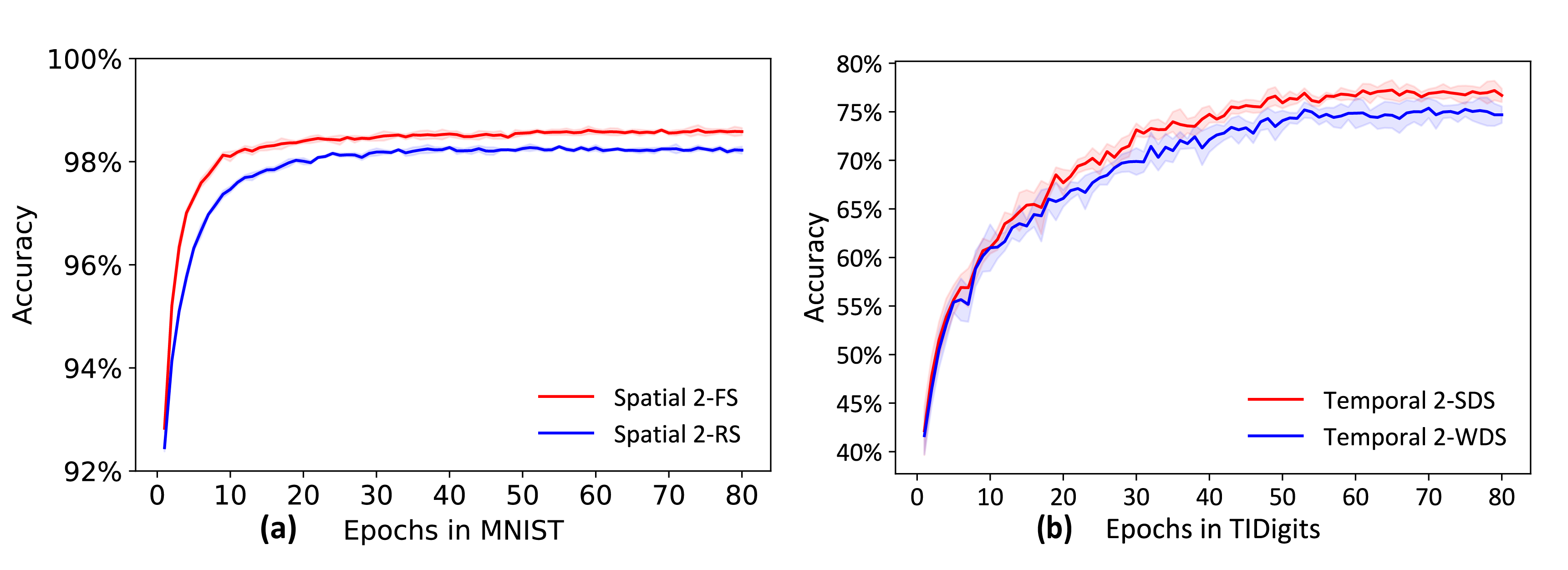}
\caption{The performance and convergence of four 2nd-order meta neurons on their source datasets.}
\label{fig_performance_meta_ori}
\end{figure}

Fig \ref{fig_dynamics}(c) showed a negative example about unreasonable meta neurons, in which $U_j(t)$ continuously decreased, and to the opposite, $V_j(t)$ reached the firing threshold quickly regardless of decay. Hence this neuron would be surprisingly non-sensitive to different neuronal inputs that contained either spatial firerate or temporal firetime information and would be discarded during following experiments.

Different with spatial meta neurons, the neuronal dynamics of temporal meta neurons were learned from temporal TIDigits dataset and shown in Fig \ref{fig_dynamics}(d,e), where the dynamic $V_j(t)$ contained more temporal information (including both spiketime and phase position). The 2nd-order Strong-Depression Spiking (2nd-SDS) neuron described a situation where the first few spikes had a strong depression for following spikes, while the 2nd-order Weak-Depression Spiking (2nd-WDS) neuron described a relatively weaker depression. These characteristics of temporal information encoding would contribute to the learning of temporal tasks. Finally dynamic parameters used in following experiments are shown in Table \ref{tab_parameter}.

Additionally, dynamic analysis was also performed on the 1st-order meta neuron in Fig. \ref{fig_dynamics}(f) in order to furtherly undersatnd the difference between the 1st-order and 2nd-order meta neurons. To some extent, the 1st-order meta neuron was similar to the spatial 2nd-FS and 2nd-RS neurons by showing relatively similar linear dynamics.

\subsection{Convergence analysis for the selected spatial and temporal meta neurons}

We tested the performance of these selected four 2nd-order meta neurons (two spatial and two temporal types) on their source datasets (MNIST and TIDigits datasets).

As shown in Fig. \ref{fig_performance_meta_ori}(a), the spatial meta neurons 2nd-FS and 2nd-RS were convergent on the spatial MNIST dataset. However, these two types of meta neurons also showed difference on their test performance, where the performance of the 2nd-FS neuron was higher than that of the 2nd-RS neuron. Similar result might also be concluded {color{red}from} temporal meta neurons, i.e., the 2nd-SDS and 2nd-WDS neurons, where both of them were convergent and the performance of the 2-SDS neuron was higher than that of the 2-WDS neuron. Hence, as a conclusion, meta neurons with faster or stronger spikes might gain a better test accuracy.

\subsection{Generalization analysis for the selected spatial and temporal meta neurons}

We chose several spatial and temporal datasets to test the performance of the selected meta neurons, including MNIST, Fashion-MNIST, NETtalk, Cifar-10 for the generalization verification of spatial neurons, and TIDigits, TIMIT, N-MNIST for that of temporal neurons.

\begin{figure*}[htpb]
\centering
\includegraphics[width=12cm]{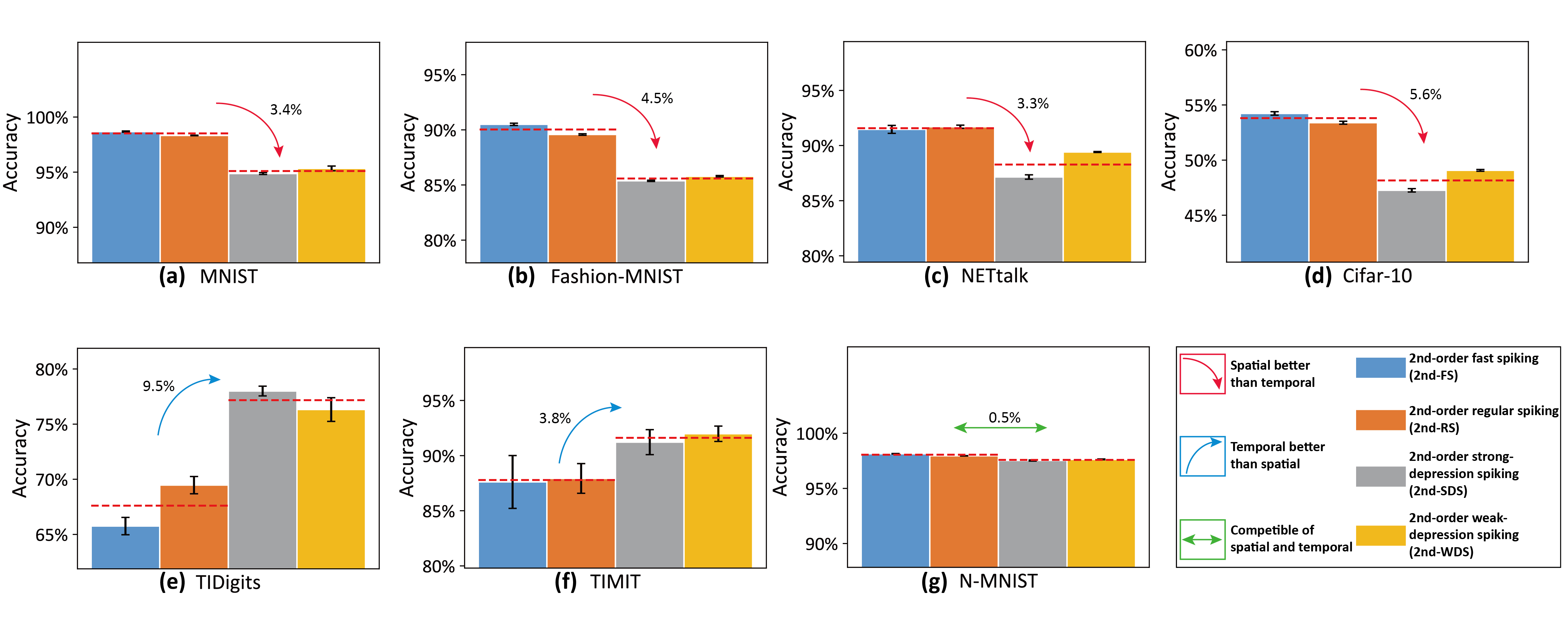}
\caption{The generalization of spatial or temporal 2nd-order meta neruons on spatial and temporal tasks.}
\label{fig_generalization_neuron}
\end{figure*}

As shown in Fig. \ref{fig_generalization_neuron}, different meta neurons showed different test accuracy while performing spatial and temporal tasks. The Fig. \ref{fig_generalization_neuron}(a,e) showed the performance of meta neurons on source datasets (datasets where their dynamic parameters were learned). As expected, the spatial meta neurons won out on their own source dataset and so were the temporal meta neurons.

For new spatial tasks, the spatial meta neurons (2nd-FS and 2nd-SDS) outpermed the temporal meta neurons (2nd-SDS and 2nd-WDS) with obviously higher mean test accuracies (the dashed red lines on the top of bars), including Fashion-MNIST (from 90.0\% to 85.5\%, accuracy), NETtalk (from 91.6\% to 88.2\%, accuracy) and Cifar-10 datasets (53.8\% to 48.1\%, accuracy), in Fig. \ref{fig_generalization_neuron}(b-d)

For new temporal tasks, including TIMIT dataset, the temporal meta neurons (2nd-SDS and 2nd-WDS) showed a relative higher mean test accuracy compared to that of spatial meta neurons, from 91.6\% to 87.7\% (accuracy), in Fig. \ref{fig_generalization_neuron}(f).

As shown in Fig. \ref{fig_generalization_neuron}(g), for new N-MNSIT dataset that contained both strong spatial and strong temporal information, the accuracy for each spatial or temporal meta neuron was similar to each other. In addition, the temporal meta neurons that fell behind spatial neurons for 3.40\% in MNIST reduced the gap to 0.5\% in N-MNIST, indicating that the temporal information might compensate temporal neurons for their weakness in handling spatial information while learning N-MNIST tasks.


\begin{table}[htbp]
    \centering
    \setlength{\tabcolsep}{0.6mm}{
    \caption{Performance of different meta neurons on chosen tasks, with unit of (\%) for following accuracies. The bold numbers are the top-2 accuracies on specific tasks.}\label{tab_result}
    \begin{tabular}{lrrrrr}
    \toprule
    Tasks  & 2nd-FS           & 2nd-RS             & 2nd-SDS       & 2nd-WDS   & 1st-order     \\
    \midrule
    MNIST         & \textbf{98.67$\pm$0.05} & 98.33$\pm$0.03 & 94.87$\pm$0.08 & 95.34$\pm$0.21 & \textbf{98.69$\pm$0.03} \\
    F-MNIST & \textbf{90.50$\pm$0.10} & 89.58$\pm$0.07 & 85.37$\pm$0.06 & 85.79$\pm$0.07 & \textbf{90.38$\pm$0.07} \\
    NETtalk       & \textbf{91.46$\pm$0.36} & \textbf{91.70$\pm$0.14} & 87.15$\pm$0.20 & 89.41$\pm$0.04 & 89.99$\pm$0.25 \\
    Cifar-10       & \textbf{54.30$\pm$0.16} & \textbf{53.39$\pm$0.13} & 47.26$\pm$0.15 & 49.07$\pm$0.08 & 52.63$\pm$0.20 \\
    TIDigits      & 65.76$\pm$0.79 & 69.47$\pm$0.78 & \textbf{78.00$\pm$0.44} & \textbf{76.32$\pm$1.08} & 61.64$\pm$1.03 \\
    TIMIT         & 87.61$\pm$2.39 & 87.93$\pm$1.35 & \textbf{91.22$\pm$1.13} & \textbf{91.98$\pm$0.70} & 76.39$\pm$2.76 \\
    N-MNIST       & \textbf{98.14$\pm$0.02} & 97.95$\pm$0.01 & 97.53$\pm$0.05 & 97.63$\pm$0.03 & \textbf{98.18$\pm$0.03} \\
    \bottomrule
    \end{tabular}}
\end{table}

Furthermore, Table \ref{tab_result} gave a summarized results of the performance of different meta neurons on different tasks. We could conclude from the table that the spatial 2nd-order spatial MDNs were more powerful on spatial tasks, while the 2nd-order temporal MDNs performed better on temporal tasks.

\subsection{The comparison of 1st- and 2nd-order neural dynamics}

\begin{figure}[htpb]
\centering
\includegraphics[width=8.8cm]{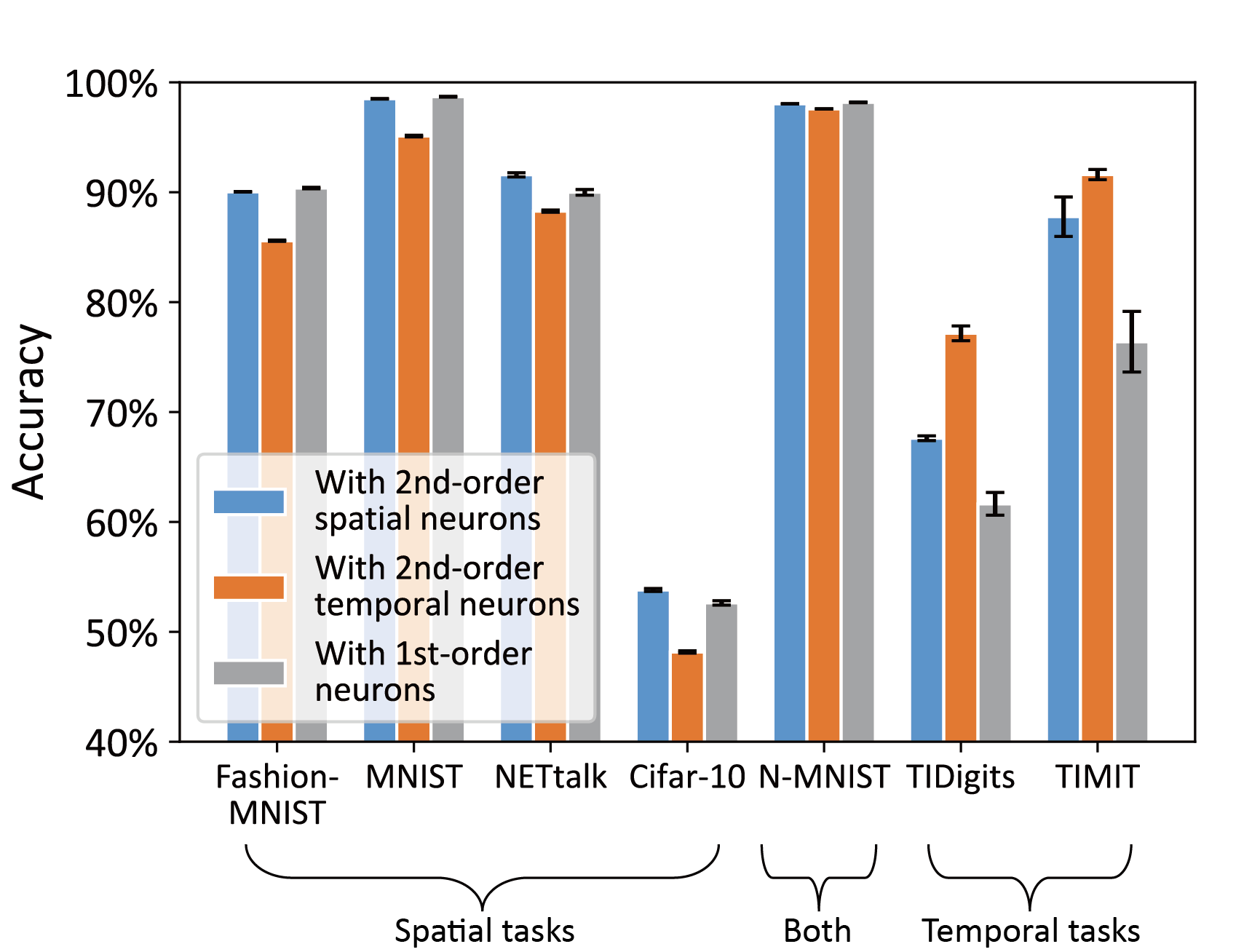}
\caption{The accuracy comparison of spatial and temporal meta neurons on different tasks.}
\label{fig_generalization_type}
\end{figure}

Additionally, we tested the peformance of 1st-order meta neuron on the same tasks and compared it with those of both 2nd-order spatial and temporal meta neurons (shown in Fig. \ref{fig_generalization_type}).

For spatial tasks, the 1st-order dynamic neuron showed similar peformance to the spatial meta neurons, and both of them exceeded the temporal meta neurons, as shown in Fig. \ref{fig_generalization_type}. For temporal tasks, the 1st-order dynamic neuron showed a worse peformance compared to the spatial meta neurons, let alone the temporal meta neurons. Such experimental results satisfied the argument that the 1st-order dynamic neuron was similar (but still less powerful) to the spatial meta neurons, suggesting the advantage of using 2nd-order dynamic neurons as meta neurons of SNNs.

\subsection{The analysis of spatio-temporal capability of different meta neurons}

The normalized performance of different meta neurons is shown in Equation (\ref{equa_cap}), 

\begin{equation}
Cap=\frac{Acc-mean(Acc)}{var(Acc)}
\label{equa_cap}
\end{equation}

where the capacity $Cap$ is calculated with the normalized value of mean $mean(Acc)$ and the standard deviation $var(Acc)$.

\begin{figure}[htpb]
\centering
\includegraphics[width=8.8cm]{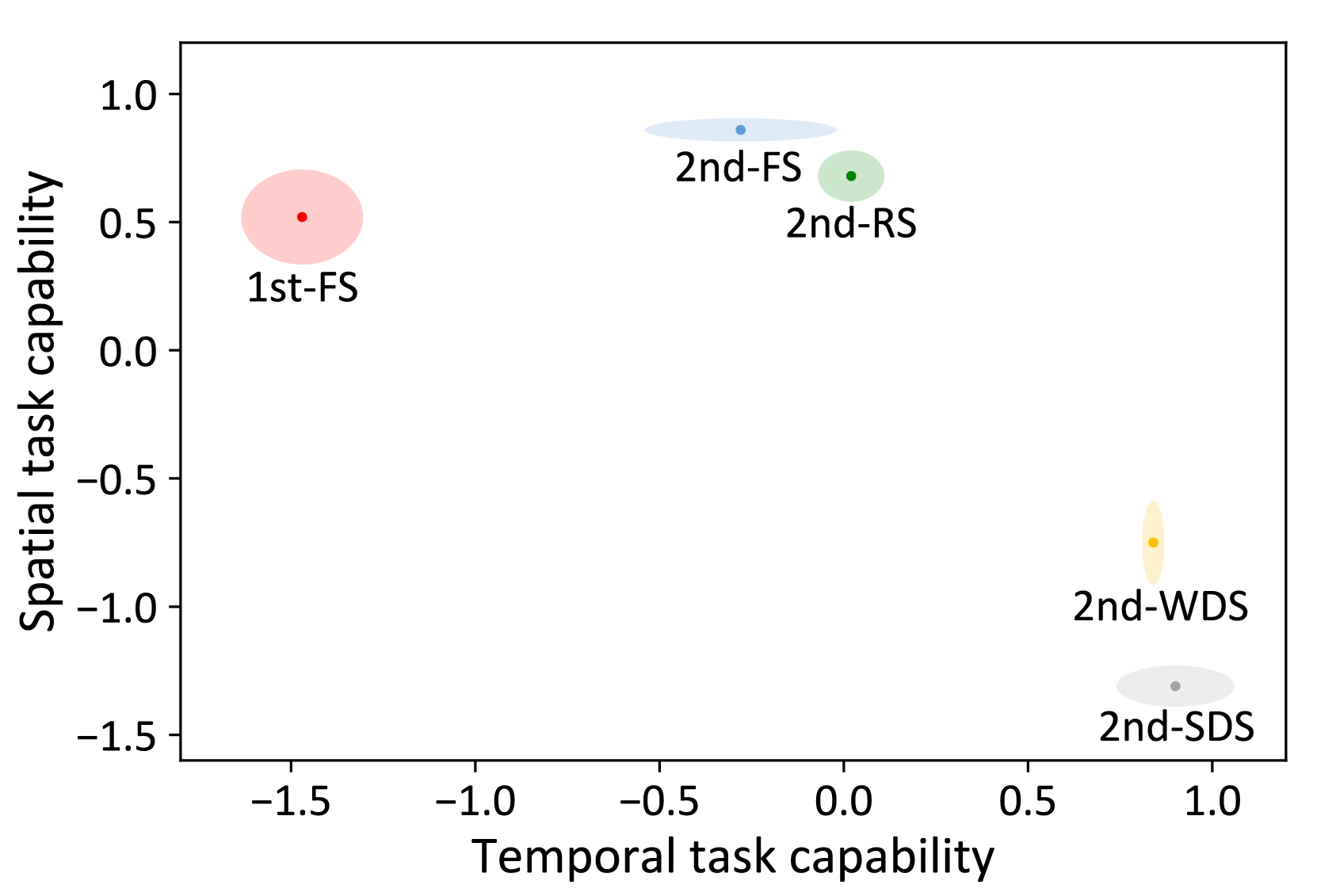}
\caption{The capability of spatial and temporal tasks for different meta neurons.}
\label{fig_capacity}
\end{figure}

As shown in Fig. \ref{fig_capacity}, different types of meta neurons showed a very different capacities. For the 1st-order fast spiking neuron, it showed strong spatial but weak temporal capacities. For the 2nd-order FS neuron, it has a stronger spatial but weaker temporal capacity compared to 2nd-order RS neuron. However, both of them were spatial-type meta neurons that showed poor capacity on temporal information processing. On the contrary, the temporal neurons, including 2nd-order WDS and SDS neurons, they showed strong temporal but weak spatial capacities. The integration of spatial and temporal meta neurons will make network strong on both spatial and temporal information, which is also one key important advantage of the biological network, where the efficient multi-type information is processed by the integration of different meta neurons.

\subsection{The comparison of our SNNs using MDNs with other SOTA SNN algorithms}

The SNNs with 1st-order or 2nd-order MDNs showed a comparable performance compared to SOTA SNN algorithms, as shwon in Table \ref{tab_SOTA}.

\begin{table}[htb]
\caption{The comparison of our MDNs improved SNNs with other SOTA SNN algorithms.}
\centering
\setlength{\tabcolsep}{0.6mm}{
\begin{tabular}{|l|r|r|r|}
\hline
Task & Algorithms & Learning rules & Accuracy \\
\hline
\multirow{3}{*}{MNIST} & SNN~\cite{RN761} & Equilibrium, STDP, 3-layers & 98.52\% \\
& SNN~\cite{RN760} & Balanced, STDP, 3-layers & 98.64\% \\
& \textbf{Ours} &  MDNs, 3-layers & \textbf{98.69\%} \\
\hline
\multirow{2}{*}{Fashion-MNIST} & SNN~\cite{RN762} & Curiosity, STDP, 3-layers & 85.74\% \\
& \textbf{Ours} & MDNs, 3-layers & \textbf{90.5\%} \\
\hline
\multirow{2}{*}{NETtalk} & SNN~\cite{RN762} & Curiosity, STDP, 3-layers & 87.20\% \\
& \textbf{Ours} & MDNs, 3-layers & \textbf{91.46\%} \\
\hline
\multirow{2}{*}{Cifar-10} & SNN\cite{RN762} & Curiosity, STDP, 3-layers & 52.85\% \\
& \textbf{Ours} & MDNs, 3-layers & \textbf{54.30\%} \\
\hline
\multirow{3}{*}{TIDigits} & SNN~\cite{wu2018a} & SOM, BP, 3-layers & \textbf{97.40\%} \\
& LSM~\cite{zhang2015a} &BP, recurrent, 3-layers& 92.30\% \\
& \textbf{Ours}  & MDNs, sequantial, 3-layers & 78.00\% \\
& \textbf{Ours}  & MDNs, non-sequantial, 3-layers & 94.00\% \\
\hline
TIMIT & \textbf{Ours} & MDNs, 3-layers & \textbf{91.98\%} \\
\hline
\multirow{2}{*}{N-MNIST} & SNN~\cite{RN828} & Convolutional 3-layers & \textbf{98.56\%} \\
& \textbf{Ours} & MDNs, 3-layers & 98.14\% \\
\hline
\end{tabular}}
\label{tab_SOTA}
\end{table}

The performance of SNNs with MDNs reached the best accuracy on spatial datasets, compared to the SOTA SNN algorithms with shallow network architectures (with 3-layers). For temporal datasets, our algorithms reached a comparable performance with other SOTA methods. The TIMIT was not compared with other algorithms for the consideration of the gender classification in our experiments, which was different from the standard TIMIT voice classification with tens of classes. The accuracy for our methods was a little lower than SOTA methods on some temporal tasks, one of the main reasons was the sequential information input instead of the whole sequence conversion as a spectrogram for the next-step classification. 

\section{Conclusion}

Until now, the research on ANNs focuses more on the complexity at the network scale, where the basic computational unit of neurons is usually designed as simple as possible, for example, some simple activation functions for non-linear spatial information conversion. This effort makes the network easily-trained with conventional backpropagation, but also leads to the exponential increasing of network scale with the growing complixity of the learning tasks. Hence the ANNs designed for relative complex tasks (e.g., ImageNet classification) are usually with a very big network size that is hard to analysis.

On the contrary, networks in biological system are different from DNNs on well balancing the complexities at both network and neuron scales. There are more types of biological neurons that are carefully designed for processing spatial or temporal information. Inspired from natural networks, here we propose Meta-Dynamic Neurons (MDNs) for a better processing of spatial and temporal informaiton in SNNs. The MDNs are designed with basic neuronal dynamics containing 1st-order and 2nd-order of membrane potentials, including the spatial and temporal meta types supported by some hyper-parameters. The MDNs generated from a spatial (MNIST) and a temporal (TIDigits) datasets first, and then extended to various other different spatio-temporal tasks (including Fashion-MNIST, NETtalk, Cifar-10, TIMIT and N-MNIST). The comparable accuracy was reached compared to other SOTA SNN algorithms, and also a better generalization was achieved by SNNs using MDNs than that without using MDNs.

We think this is a small step towards the biologically-plausible efficient information processing from the perspective of meta neurons. Next, a further research on meta circuits (or network Motifs) will also contribute this areas towards a better understanding of the strategy of information processing in the biological brain.

\section*{Acknowledgment}
This paper is supported by the National Natural Science Foundation of China (No. 61806195), the Strategic Priority Research Program of the Chinese Academy of Sciences (No. XDBS01070000), and the Beijing Brain Science Project (No. Z181100001518006). The source code link on the github page is https://github.com/thomasaimondy/MDN-SNN.

\section*{Author contributions}
X.C., T.Z. and B.X. designed the study. T.Z., X.C. and S.J. performed the experiments and analyses. They wrote the paper together.

\section*{Competing interests}
The authors declare no competing interests.

\bibliography{thomas}

\begin{thebibliography}{10}
\expandafter\ifx\csname url\endcsname\relax
  \def\url#1{\texttt{#1}}\fi
\expandafter\ifx\csname urlprefix\endcsname\relax\def\urlprefix{URL }\fi
\expandafter\ifx\csname href\endcsname\relax
  \def\href#1#2{#2} \def\path#1{#1}\fi

\bibitem{RN699}
D.~Hassabis, D.~Kumaran, C.~Summerfield, M.~Botvinick, Neuroscience-inspired
  artificial intelligence, Neuron 95~(2) (2017) 245--258.

\bibitem{RN668}
A.~H. Marblestone, G.~Wayne, K.~P. Kording, Toward an integration of deep
  learning and neuroscience, Frontiers in Computational Neuroscience 10 (2016)
  94.

\bibitem{RN740}
A.~Nguyen, J.~Yosinski, J.~Clune, Deep neural networks are easily fooled: High
  confidence predictions for unrecognizable images, in: Computer Vision and
  Pattern Recognition, 2015, pp. 427--436.

\bibitem{RN789}
X.~Chen, Y.~Duan, R.~Houthooft, J.~Schulman, I.~Sutskever, P.~Abbeel, Infogan:
  interpretable representation learning by information maximizing generative
  adversarial nets, in: Advances in Neural Information Processing Systems,
  2016, pp. 2180--2188.

\bibitem{RN660}
Y.~LeCun, Y.~Bengio, G.~Hinton, Deep learning, Nature 521~(7553) (2015)
  436--444.

\bibitem{RN726}
T.~P. Lillicrap, A.~Santoro, L.~Marris, C.~J. Akerman, G.~Hinton,
  Backpropagation and the brain, Nature Reviews Neuroscience (2020) 1--12.

\bibitem{RN405}
S.~Sabour, N.~Frosst, G.~E. Hinton, Dynamic routing between capsules, in:
  Advances in Neural Information Processing Systems, 2017, pp. 3859--3869.

\bibitem{RN378}
I.~D. Alex~Graves, Greg~Wayne, Google mind - neural turing machines.

\bibitem{RN788}
J.~Kirkpatrick, R.~Pascanu, N.~C. Rabinowitz, J.~Veness, G.~Desjardins, A.~A.
  Rusu, K.~Milan, J.~Quan, T.~Ramalho, A.~Grabskabarwinska, Overcoming
  catastrophic forgetting in neural networks, Proceedings of the National
  Academy of Sciences of the United States of America 114~(13) (2017)
  3521--3526.

\bibitem{RN515}
A.~Fire, S.-C. Zhu, Inferring hidden statuses and actions in video by causal
  reasoning, in: Proceedings of the IEEE Conference on Computer Vision and
  Pattern Recognition Workshops, 2017, pp. 48--56.

\bibitem{RN691}
Y.~Bengio, T.~Mesnard, A.~Fischer, S.~Zhang, Y.~Wu, Stdp as presynaptic
  activity times rate of change of postsynaptic activity approximates
  back-propagation, Neural Computation 10.

\bibitem{RN674}
Y.~Dan, M.-m. Poo, Spike timing-dependent plasticity of neural circuits, Neuron
  44~(1) (2004) 23--30.

\bibitem{RN693}
R.~S. Zucker, Short-term synaptic plasticity, Annual Review of Neuroscience
  12~(1) (1989) 13--31.

\bibitem{RN684}
T.~J. Teyler, P.~DiScenna, Long-term potentiation, Annual Review of
  Neuroscience 10~(1) (1987) 131--161.

\bibitem{RN701}
M.~Ito, Long-term depression, Annual Review of Neuroscience 12~(1) (1989)
  85--102.

\bibitem{RN468}
P.~U. Diehl, M.~Cook, Unsupervised learning of digit recognition using
  spike-timing-dependent plasticity, Front Comput Neurosci 9 (2015) 99.
\newblock \href {http://dx.doi.org/10.3389/fncom.2015.00099}
  {\path{doi:10.3389/fncom.2015.00099}}.

\bibitem{RN704}
L.~F. Abbott, B.~DePasquale, R.-M. Memmesheimer, Building functional networks
  of spiking model neurons, Nature Neuroscience 19~(3) (2016) 350.

\bibitem{RN679}
W.~Maass, Networks of spiking neurons: the third generation of neural network
  models, Neural Networks 10~(9) (1997) 1659--1671.

\bibitem{RN682}
L.~F. Abbott, S.~B. Nelson, Synaptic plasticity: taming the beast, Nature
  Neuroscience 3~(11s) (2000) 1178.

\bibitem{RN618}
H.~Zeng, J.~R. Sanes, Neuronal cell-type classification: challenges,
  opportunities and the path forward, Nature Reviews Neuroscience 18~(9) (2017)
  530.

\bibitem{RN588}
D.~Arendt, J.~M. Musser, C.~V. Baker, A.~Bergman, C.~Cepko, D.~H. Erwin,
  M.~Pavlicev, G.~Schlosser, S.~Widder, M.~D. Laubichler, The origin and
  evolution of cell types, Nature Reviews Genetics 17~(12) (2016) 744.

\bibitem{RN745}
E.~M. Izhikevich, Simple model of spiking neurons, IEEE Trans Neural Netw
  14~(6) (2003) 1569--72.
\newblock \href {http://dx.doi.org/10.1109/TNN.2003.820440}
  {\path{doi:10.1109/TNN.2003.820440}}.

\bibitem{RN426}
Auto machine learning, methods, systems, challenges.

\bibitem{RN732}
S.~Woźniak, A.~Pantazi, T.~Bohnstingl, E.~Eleftheriou, Deep learning
  incorporating biologically inspired neural dynamics and in-memory computing,
  Nature Machine Intelligence 2~(6) (2020) 325--336.
\newblock \href {http://dx.doi.org/10.1038/s42256-020-0187-0}
  {\path{doi:10.1038/s42256-020-0187-0}}.

\bibitem{RN819}
A.~L. Hodgkin, A.~F. Huxley, A quantitative description of membrane current and
  its application to conduction and excitation in nerve, J Physiol 117~(4)
  (1952) 500--44.
\newblock \href {http://dx.doi.org/10.1113/jphysiol.1952.sp004764}
  {\path{doi:10.1113/jphysiol.1952.sp004764}}.

\bibitem{RN501}
F.~Zenke, E.~J. Agnes, W.~Gerstner, Diverse synaptic plasticity mechanisms
  orchestrated to form and retrieve memories in spiking neural networks, Nature
  communications 6.

\bibitem{RN697}
A.~Alemi, C.~K. Machens, S.~Deneve, J.-J. Slotine, Learning nonlinear dynamics
  in efficient, balanced spiking networks using local plasticity rules, in:
  Thirty-Second AAAI Conference on Artificial Intelligence, 2018.

\bibitem{RN725}
M.~Shi, T.~Zhang, Y.~Zeng, A curiosity-based learning method for spiking neural
  networks, Frontiers in Computational Neuroscience 14 (2020) 7.

\bibitem{RN705}
F.~Zenke, E.~J. Agnes, W.~Gerstner, Diverse synaptic plasticity mechanisms
  orchestrated to form and retrieve memories in spiking neural networks, Nature
  Communications 6 (2015) 6922.

\bibitem{RN689}
S.~B. Shrestha, G.~Orchard, SLAYER: Spike layer error reassignment in time,
  Curran Associates, Inc., 2018, pp. 1412--1421.

\bibitem{RN711}
Y.~Zeng, T.~Zhang, B.~Xu, Improving multi-layer spiking neural networks by
  incorporating brain-inspired rules, Science China Information Sciences 60~(5)
  (2017) 052201.

\bibitem{RN723}
T.~Zhang, Y.~Zeng, D.~Zhao, M.~Shi, A plasticity-centric approach to train the
  non-differential spiking neural networks, in: Thirty-Second AAAI Conference
  on Artificial Intelligence.

\bibitem{RN820}
W.~Gerstner, R.~Ritz, J.~L. van Hemmen, Why spikes? hebbian learning and
  retrieval of time-resolved excitation patterns, Biological Cybernetics
  69~(5-6) (1993) 503--515.
\newblock \href {http://dx.doi.org/10.1007/bf00199450}
  {\path{doi:10.1007/bf00199450}}.

\bibitem{RN821}
Y.~C. Yoon, Lif and simplified srm neurons encode signals into spikes via a
  form of asynchronous pulse sigma-delta modulation, IEEE Trans Neural Netw
  Learn Syst 28~(5) (2017) 1192--1205.
\newblock \href {http://dx.doi.org/10.1109/TNNLS.2016.2526029}
  {\path{doi:10.1109/TNNLS.2016.2526029}}.

\bibitem{RN237}
E.~M. Izhikevich, Which model to use for cortical spiking neurons?, IEEE
  transactions on neural networks 15~(5) (2004) 1063--1070.

\bibitem{RN696}
S.~R. Kheradpisheh, M.~Ganjtabesh, S.~J. Thorpe, e.~Masquelier, Timoth\'e,
  Stdp-based spiking deep convolutional neural networks for object recognition,
  Neural Networks 99 (2018) 56--67.

\bibitem{RN641}
A.~Tavanaei, A.~Maida, Bp-stdp: Approximating backpropagation using spike
  timing dependent plasticity, Neurocomputing 330 (2019) 39--47.

\bibitem{RN694}
M.~Mozafari, M.~Ganjtabesh, A.~Nowzari-Dalini, S.~J. Thorpe, e.~Masquelier,
  Timoth\'e, Bio-inspired digit recognition using reward-modulated
  spike-timing-dependent plasticity in deep convolutional networks, Pattern
  Recognition 94 (2019) 87--95.

\bibitem{RN673}
T.~Zhang, Y.~Zeng, D.~Zhao, B.~Xu, Brain-inspired balanced tuning for spiking
  neural networks, in: International Joint Conferences on Artificial
  Intelligence, 2018, pp. 1653--1659.

\bibitem{RN643}
Y.~Wu, L.~Deng, G.~Li, J.~Zhu, L.~Shi, Spatio-temporal backpropagation for
  training high-performance spiking neural networks, Frontiers in neuroscience
  12.

\bibitem{RN686}
J.~H. Lee, T.~Delbruck, M.~Pfeiffer, Training deep spiking neural networks
  using backpropagation, Frontiers in Neuroscience 10.

\bibitem{diehl2015fast}
P.~U. Diehl, D.~Neil, J.~Binas, M.~Cook, S.-C. Liu, M.~Pfeiffer,
  Fast-classifying, high-accuracy spiking deep networks through weight and
  threshold balancing, in: Proceddings of the 2015 International Joint
  Conference on Neural Networks, IEEE, 2015, pp. 1--8.

\bibitem{RN680}
F.~Zenke, S.~Ganguli, Superspike: Supervised learning in multilayer spiking
  neural networks, Neural Computation 30~(6) (2018) 1514--1541.

\bibitem{zhang2015a}
Y.~Zhang, P.~Li, Y.~Jin, Y.~Choe, A digital liquid state machine with
  biologically inspired learning and its application to speech recognition,
  IEEE Transactions on Neural Networks 26~(11) (2015) 2635--2649.

\bibitem{RN822}
G.~Bellec, F.~Scherr, A.~Subramoney, E.~Hajek, D.~Salaj, R.~Legenstein,
  W.~Maass, A solution to the learning dilemma for recurrent networks of
  spiking neurons, Nature Communications 11~(1).
\newblock \href {http://dx.doi.org/10.1038/s41467-020-17236-y}
  {\path{doi:10.1038/s41467-020-17236-y}}.

\bibitem{wu2018a}
J.~Wu, Y.~Chua, M.~Zhang, H.~Li, K.~C. Tan, A spiking neural network framework
  for robust sound classification, Frontiers in Neuroscience 12.

\bibitem{RN762}
M.~Shi, T.~Zhang, Y.~Zeng, A curiosity-based learning method for spiking neural
  networks, Frontiers in Computational Neuroscience 14 (2020) 7.

\bibitem{RN657}
Y.~LeCun, The mnist database of handwritten digits,
  http://yann.lecun.com/exdb/mnist/.

\bibitem{RN212}
T.~J. Sejnowski, C.~R. Rosenberg, NETtalk: A parallel network that learns to
  read aloud, MIT Press, 1988.

\bibitem{RN800}
A.~Krizhevsky, G.~Hinton, Learning multiple layers of features from tiny
  images, Tech Report.

\bibitem{orchard2015converting}
G.~Orchard, A.~Jayawant, G.~Cohen, N.~V. Thakor, Converting static image
  datasets to spiking neuromorphic datasets using saccades, Frontiers in
  Neuroscience 9 (2015) 437--437.

\bibitem{RN798}
R.~G. Leonard, G.~Doddington, Tidigits ldc93s10, Web Download. Philadelphia:
  Linguistic Data Consortium.

\bibitem{RN823}
J.~S. Garofolo, Timit acoustic phonetic continuous speech corpus, Linguistic
  Data Consortium, 1993.

\bibitem{RN761}
T.~Zhang, Y.~Zeng, M.~Shi, D.~Zhao, A plasticity-centric approach to train the
  non-differential spiking neural networks, in: Thirty-Second AAAI Conference
  on Artificial Intelligence, 2018, pp. 620--628.

\bibitem{RN760}
T.~Zhang, Y.~Zeng, D.~Zhao, B.~Xu, Brain-inspired balanced tuning for spiking
  neural networks, in: International Joint Conference on Artificial
  Intelligence, 2018, pp. 1653--1659.

\bibitem{RN828}
J.~H. Lee, T.~Delbruck, M.~Pfeiffer, Training deep spiking neural networks
  using backpropagation, Front Neurosci 10 (2016) 508.
\newblock \href {http://dx.doi.org/10.3389/fnins.2016.00508}
  {\path{doi:10.3389/fnins.2016.00508}}.

\end{thebibliography}

\end{document}